\documentclass[10pt,twocolumn,letterpaper]{article}

\usepackage{iccv}
\usepackage{times}
\usepackage{epsfig}
\usepackage{graphicx}
\usepackage{amsmath}
\usepackage{amssymb}
\usepackage{multirow}
\usepackage{array}

\usepackage[breaklinks=true,bookmarks=false]{hyperref}
\iccvfinalcopy 
\def\iccvPaperID{****} 

\begin{document}

\title{Pose-driven Deep Convolutional Model for Person Re-identification}

\author{Chi Su\textsuperscript{1}\footnotemark[1]~\footnotemark[2]~, Jianing Li\textsuperscript{1}\footnotemark[1]~, Shiliang Zhang\textsuperscript{1}, Junliang Xing\textsuperscript{2}, Wen Gao\textsuperscript{1}, Qi Tian\textsuperscript{3}\\
\normalsize{\textsuperscript{1}School of Electronics Engineering and Computer Science, Peking University, Beijing 100871, China}\\
\normalsize{\textsuperscript{2}National Laboratory of Pattern Recognition, Institute of Automation, Chinese Academy of Sciences, Beijing 100190, China}\\
\normalsize{\textsuperscript{3}Department of Computer Science, University of Texas at San Antonio, San Antonio, TX 78249-1604, USA}\\
{\tt\small suchi@kingsoft.com,kaneiri1024@gmail.com,\{slzhang.jdl,wgao\}@pku.edu.cn,}\\
{\tt\small jlxing@nlpr.ia.ac.cn,qi.tian@utsa.edu}
}

\renewcommand{\thefootnote}{\fnsymbol{footnote}}
\maketitle
\footnotetext[1]{indicates equal contribution.}
\footnotetext[2]{Chi Su finished this work when he was a Ph.d candiadate in Peking University, now he has got his Ph.d degree and is working in Beijing Kingsoft Cloud Network Technology Co.,Ltd, No.33,xiaoying Rd.W., HaiDian Dist., Beijing 100085, China}
\begin{abstract}
Feature extraction and matching are two crucial components in person Re-Identification (ReID). The large pose deformations and the complex view variations exhibited by the captured person images significantly increase the difficulty of learning and matching of the features from person images. To overcome these difficulties, in this work we propose a Pose-driven Deep Convolutional (PDC) model to learn improved feature extraction and matching models from end to end. Our deep architecture explicitly leverages the human part cues to alleviate the pose variations and learn robust feature representations from both the global image and different local parts. To match the features from global human body and local body parts, a pose driven feature weighting sub-network is further designed to learn adaptive feature fusions. Extensive experimental analyses and results on three popular datasets demonstrate significant performance improvements of our model over all published state-of-the-art methods.
\end{abstract}

\section{Introduction}
\label{sec1}
Person Re-Identification (ReID) is an important component in a video surveillance system. Here person ReID refers to the process of identifying a probe person from a gallery captured by different cameras, and is generally deployed in the following scenario: given a probe image or video sequence containing a specific person under a certain camera, querying the images, locations, and time stamps of this person from other cameras.

Despite decades of studies, the person ReID problem is still far from being solved. This is mainly because of challenging situations like complex view variations and large pose deformations on the captured person images. Most of traditional works try to address these challenges with the following two approaches: (1) representing the visual appearance of a person using customized local invariant features extracted from images~\cite{farenzena2010person, cheng2011custom, ma2012bicov, liu2012person, zhao2013unsupervised, wang2014person, Zheng_2015_CVPR, su2017attributes} or (2) learning a discriminative distance metric to reduce the distance among features of images containing the same person~\cite{ma2013domain, dikmen2011pedestrian, hirzer2012relaxed, pedagadi2013local, yan2007graph, kostinger2012large, xiong2014person, liu2013pop, li2013learning, zheng2013re, wang2014personvideo, chen2015similarity, liao2015person, chen2015mirror, shen2015person, liao2015efficient, ding2015deep, pengunsupervised, zhang2016learning}. Because the human poses and viewpoints are uncontrollable in real scenarios, hand-coded features may be not robust enough to pose and viewpoint variations. Distance metric is computed for each pair of cameras, making distance metric learning based person ReID suffers from the $\mathcal{O}^2$ computational complexity.
\begin{figure}
\centering 
\includegraphics[width=0.99\linewidth]{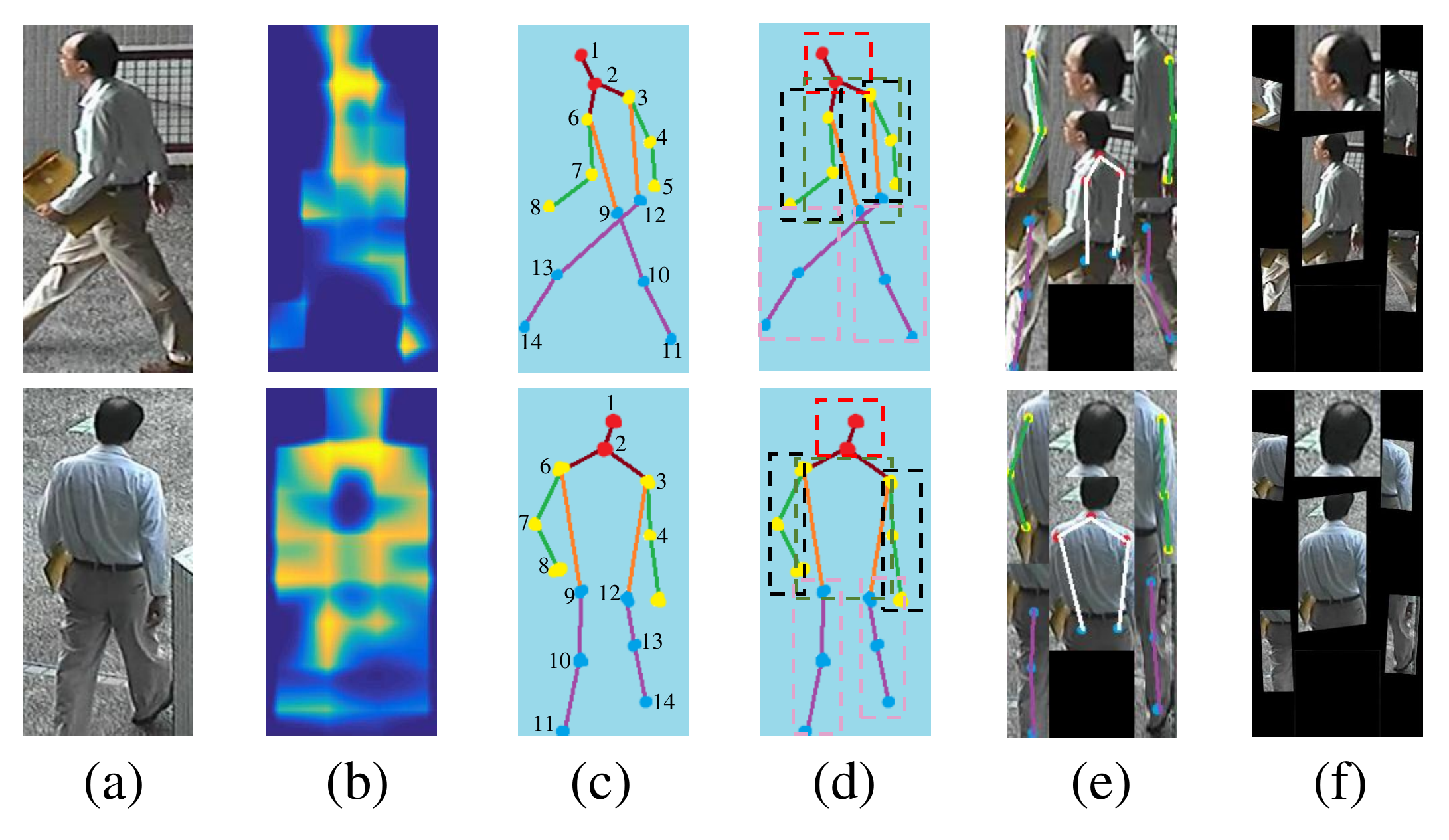}\\
\vspace{-2mm}
\caption{Illustration of part extraction and pose normalization in our Feature Embedding sub-Net (FEN). Response maps of 14 body joints (b) are first generated from the original image in (a). 14 body joints in (c) and 6 body parts in (d) can hence be inferred. The part regions are firstly rotated and resized in (e), then normalized by Pose Transform Network in (f). }
\vspace{-4mm}
\label{fig:pdfes-example}
\end{figure}

In recent years, deep learning has demonstrated strong model capabilities and obtains very promising performances in many computer vision tasks \cite{Krizhevsky:NIPS12, Girshick:PAMI15, long2015fully, CVPR15FaceNet, deng2013fine}. Meanwhile, the release of person ReID datasets like CUHK 03~\cite{li2014deepreid}, Market-1501~\cite{zheng2015scalable}, and MARS~\cite{zheng2016mars}, both of which contain many annotated person images, makes training deep models for person ReID feasible. Therefore, many researchers attempt to leverage deep models in person ReID \cite{ahmed5improved, ding2015deep, xiao2016learning, varior2016gated, su2016deep, zheng2016mars, geng2016deep, yaolarge, su2017multiType, yao2017deep}. Most of these methods first learn a pedestrian feature and then compute Euclidean distance to measure the similarity between two samples. More specifically, existing deep learning based person ReID approaches can be summarized into two categories: 1) use Softmax Loss with person ID labels to learn a global representation \cite{ahmed5improved, ding2015deep, xiao2016learning, varior2016gated, su2016deep, zheng2016mars, geng2016deep}, and 2) first learn local representations using predefined rigid body parts, then fuse the local and global representations \cite{cheng2016person, varior2016siamese, shi2016embedding} to depict person images. Deep learning based methods have demonstrated significant performance improvements over the traditional methods. Although these approaches have achieved remarkable results on mainstream person ReID datasets, most of them do not consider pose variation of human body.

Because pose variations may significantly change the appearance of a person, considering the human pose cues is potential to help person re-identification. Although there are several methods~\cite{cheng2016person, varior2016siamese, shi2016embedding} that segment the person images according to the predefined configuration, such simple segmentation can not capture the pose cues effectively. Some recent works~\cite{zheng2017pose, zhao2017spindle} attempt to use pose estimation algorithms to predict human pose and then train deep models for person ReID. However, they use manually cropped human body parts and their models are not trained from end to end. Therefore, the potential of pose information to boost the ReID performance has not been fully explored.

To better alleviate the challenges from pose variations, we propose a Pose-driven Deep Convolutional (PDC) model for person ReID. The proposed PDC model learns the global representation depicting the whole body and local representations depicting body parts simultaneously. The global representation is learned using the Softmax Loss with person ID labels on the whole input image. For the learning of local representations, a novel Feature Embedding sub-Net (FEN) is proposed to learn and readjust human parts so that parts are affine transformed and re-located at more reasonable regions which can be easily recognizable through two different cameras. In Feature Embedding sub-Net, each body part region is first automatically cropped. The cropped part regions are hence transformed by a Pose Transformation Network (PTN) to eliminate the pose variations. The local representations are hence learned on the transformed regions. We further propose a Feature Weighting sub-Net (FWN) to learn the weights of global representations and local representations on different parts. Therefore, more reasonable feature fusion is conducted to facilitate feature similarity measurement.

\begin{figure}
\centering 
\includegraphics[width=1\linewidth]{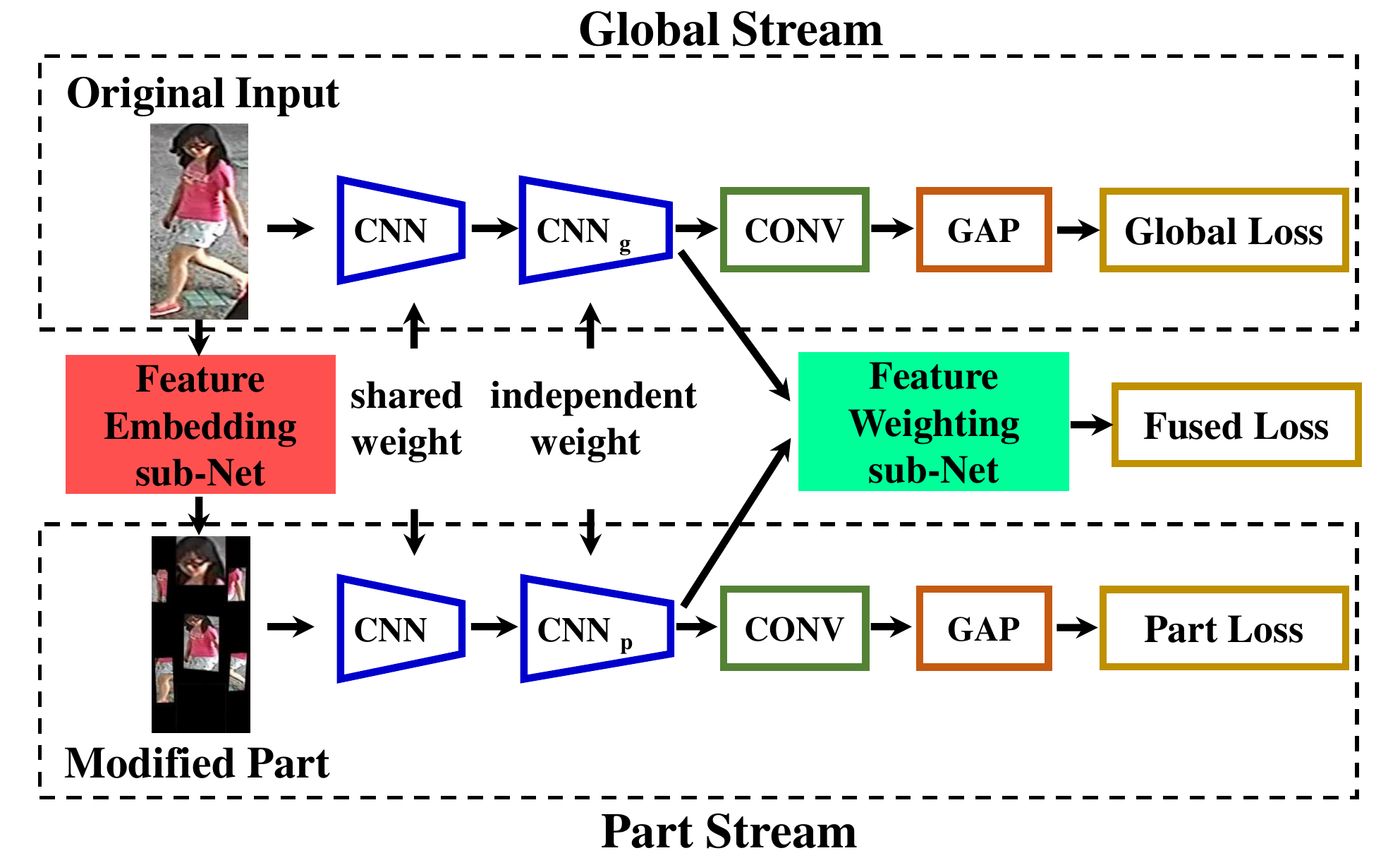}\\
\vspace{-1mm}
\caption{Flowchart of Pose-driven Deep Convolutional (PDC) model. Feature Embedding sub-Net (FEN) leverages human pose information and transforms a global body image into an image containing normalized part regions. Feature Weighting sub-Net (FWN) automatically learns the weights of the different part representations to facilitate feature similarity measurement.}
\vspace{-4mm}
\label{fig:framewrok}
\end{figure}

Some more detailed descriptions to our local representation generation are illustrated in Fig.\ref{fig:pdfes-example}. Our method first locates the key body joints from the input image, \eg, illustrated in Fig.\ref{fig:pdfes-example}~(c). From the detected joints, six body parts are extracted, \eg, shown in Fig.\ref{fig:pdfes-example}(d). As shown in Fig.\ref{fig:pdfes-example}(e), those parts are extracted and normalized into fixed sizes and orientations. Finally, they are fed into the Pose Transformation Network (PTN) to further eliminate the pose variations. With the normalized and transformed part regions, \eg, Fig.\ref{fig:pdfes-example}~(f), local representations are learned by training the deep neural network. Different parts commonly convey different levels of discriminative cues to identify the person. We thus further learn weights for representations on different parts with a sub-network.

Most of current deep learning based person ReID works do not consider the human pose cues and the weights of representation on different parts. This paper proposes a novel deep architecture that transforms body parts into normalized and homologous feature representations to better overcome the pose variations. Moreover, a sub-network is proposed to automatically learn weights for different parts to facilitate feature similarity measurement. Both the representation and weighting are learned jointly from end to end. Since pose estimation is not the focus of this paper, the used pose estimation algorithm, \ie, Fully Convolutional Networks(FCN)~\cite{long2015fully} based pose estimation method is simple and trained independently. Once the FCN is trained, it is incorporated in our framework, which is hence trained in an end-to-end manner, \ie, using images as inputs and person ID labels as outputs. Experimental results on three popular datasets show that our algorithm significantly outperforms many state-of-the-art ones.

\begin{table*}[htbp]
\footnotesize
\tabcolsep=5pt
\caption{Detailed structure of the proposed Pose-driven Deep Convolutional (PDC) model.}
  \centering
    \begin{tabular}{l|c|c|c|c|c|c|c|c|c|c}
    \hline \multirow{2}{*}{type}&share& {patch size} & \multirow{2}{*}{output size} & \multirow{2}{*}{depth} & \multirow{2}{*}{\#1$\times$1} & {\#3$\times$3} & \multirow{2}{*}{\#3$\times$3} & {double\#3$\times$3} & {double} & \multirow{2}{*}{pool proj} \\
    &weight&/stride&&&&reduce&&reduce&\#3$\times$3&\\
    \hline\hline {data} &-& {-} & {512$\times$ 256$\times$ 3} & {-} & {-} & {-} & {-} & {-} & {-} & {-} \\
    \hline {convolution} &Yes& {7$\times$ 7/2} & {256$\times$ 128$\times$ 64} & {1} & {-} & {-} & {-} & {-} & {-} & {-} \\
    \hline {max pool} &-& {3$\times$ 3/2} & {128$\times$ 64$\times$ 64} & {0} & {-} & {-} & {-} & {-} & {-} & {-} \\
    \hline {convolution} &Yes& {3$\times$ 3/1} & {128$\times$ 64$\times$ 192} & {1} & {-} & {64} & {192} & {-} & {-} & {-} \\
    \hline {max pool} &-& {3$\times$ 3/2} & {64$\times$ 32$\times$ 192} & {0} & {-} & {-} & {-} & {-} & {-} & {-} \\
    \hline {inception(3a)} &Yes&{-} & {64$\times$ 32$\times$ 256} & {3} & {64} & {64} & {64} & {64} & {96} & {avg+32} \\
    \hline {inception(3b)} &Yes&{-} & {64$\times$ 32$\times$ 320} & {3} & {64} & {64} & {96} & {64} & {96} & {avg+64} \\
    \hline {inception(3c)} &Yes&{stride 2} & {32$\times$ 16$\times$ 576} & {3} & {0} & {128} & {160} & {64} & {96} & {max+pass through} \\
    \hline {inception(4a)} &Yes&{-} & {32$\times$ 16$\times$ 576} & {3} & {224} & {64} & {96} & {96} & {128} & {avg+128} \\
    \hline {inception(4b)} &Yes&{-} & {32$\times$ 16$\times$ 576} & {3} & {192} & {96} & {128} & {96} & {128} & {avg+128} \\
    \hline {inception(4c)} &Yes&{-} & {32$\times$ 16$\times$ 576} & {3} & {160} & {128} & {160} & {128} & {160} & {avg+128} \\
    \hline {inception(4d)} &Yes&{-} & {32$\times$ 16$\times$ 576} & {3} & {96} & {128} & {192} & {160} & {192} & {avg+128} \\
    \hline {inception(4e)} &Yes&{stride 2} & {16$\times$ 8$\times$ 1024} & {3} & {0} & {128} & {192} & {192} & {256} & {max+pass through} \\
    \hline {inception(5a)} &No&{-} & {16$\times$ 8$\times$ 1024} & {3} & {352} & {192} & {320} & {160} & {224} & {avg+128} \\
    \hline {inception(5b)} &No&{-} & {16$\times$ 8$\times$ 1024} & {3} & {352} & {192} & {320} & {192} & {224} & {max+128} \\
    \hline {convolution} &No&{1$\times$ 1/1} & {16$\times$ 8$\times$ class num} & {1} & {-} & {-} & {-} & {-} & {-} & {-} \\
    \hline {ave pool} &-&{global pooling} & {1$\times$ 1$\times$ class num} & {0} & {-} & {-} & {-} & {-} & {-} & {-} \\
    \hline
    \end{tabular}%
    \vspace{-4mm}
  \label{table:baseline}%
\end{table*}%

\section{Related Work}
Traditional algorithms perform person re-identification through two ways: (a) acquiring robust local features visually representing a person's appearance and then encoding them ~\cite{farenzena2010person,cheng2011custom,ma2012bicov,liu2012person,zhao2013unsupervised,wang2014person,Zheng_2015_CVPR}; (b) closing the gap between a person's different features by learning a discriminative distance metric~\cite{ma2013domain,dikmen2011pedestrian,hirzer2012relaxed,pedagadi2013local,yan2007graph,kostinger2012large,xiong2014person,liu2013pop,li2013learning,zheng2013re,wang2014personvideo,chen2015similarity,liao2015person,chen2015mirror,shen2015person,liao2015efficient,ding2015deep,pengunsupervised,zhang2016learning}. Some recent works~\cite{ahmed5improved,ding2015deep,xiao2016learning,varior2016gated,su2016deep,zheng2016mars,geng2016deep,cheng2016person,varior2016siamese,shi2016embedding,zheng2017pose,zhao2017spindle} have started to apply deep learning in person ReID and achieved promising performance. In the following, we briefly review recent deep learning based person ReID methods.

Deep learning is commonly used to either learn a person's representation or the distance metric. When handling a pair of person images, existing deep learning methods usually learn feature representations of each person by using a deep matching function from convolutional features~\cite{ahmed5improved,li2014deepreid,xiao2016learning,geng2016deep} or from the Fully Connected (FC) features ~\cite{yi2014deep,shi2016embedding,zheng2016mars}. Apart from deep metric learning methods, some algorithms first learn image representations directly with the Triplet Loss or the Siamese Contrastive Loss, then utilize Euclidean distance for comparison~\cite{wangjoint,cheng2016person,ding2015deep,varior2016gated}. Wang \etal~\cite{wangjoint} use a joint learning framework to unify single-image representation and cross-image representation using a doublet or triplet CNN. Shi \etal~\cite{shi2016embedding} propose a moderate positive mining method to use
deep distance metric learning for person ReID. Another novel method~\cite{shi2016embedding} learns deep attributes feature for ReID with semi-supervised learning. Xiao \etal~\cite{xiao2016learning} train one network with several person ReID datasets using a Domain Guided Dropout algorithm.

Predefined rigid body parts are also used by many deep learning based methods~\cite{cheng2016person,varior2016siamese,shi2016embedding} for the purpose of learning local pedestrian features. Different from these algorithms, our work and the ones in~\cite{zheng2017pose,zhao2017spindle} use more accurate human pose estimation algorithms to acquire human pose features. However, due to the limited accuracy of pose estimation algorithms as well as reasons like occlusion and lighting change, pose estimation might be not accurate enough. Moreover, different parts convey different levels of discriminative cues. Therefore, we normalize the part regions to get more robust feature representation using Feature Embedding sub-Net (FEN) and propose a Feature Weighting sub-Net (FWN) to learn the weight for each part feature. In this way, the part with high discriminative power can be identified and emphasized. This also makes our work different from existing ones~\cite{zheng2017pose,zhao2017spindle}, which do not consider the inaccuracy of human poses estimation and weighting on different parts features.

\section{Pose-driven Deep ReID Model }
In this section, we describe the overall framework of the proposed approach, where we mainly introduce the Feature Embedding sub-Net (FEN) and the Feature Weighting sub-Net (FWN). Details about the training and test procedures of the proposed approach will also be presented.

\subsection{Framework}
Fig.\ref{fig:framewrok} shows the framework of our proposed deep ReID model. It can be seen that the global image and part images are simultaneously considered during each round of training. Given a training sample, we use an human pose estimation algorithm to acquire the locations of human pose joints. These pose joints are combined into different human body parts. The part regions are first transformed using our Feature Embedding sub-Net (FEN) and then are combined to form a new modified part image containing the normalized body parts. The global image and the new modified part image are then fed into our CNN together. The two images share the same weights for the first several layers, then have their own network weights in the subsequent layers. At last, we use Feature Weighting sub-Net (FWN) to learn the weights of part features before fusing them with the global features for final Softmax Loss computation.

Considering that pedestrian images form different datasets have different sizes, it is not appropriate to directly use the CNN models pre-trained on the ImageNet dataset ~\cite{imagenet_cvpr09}. We thus modify and design a network based on the GoogLeNet~\cite{szegedy2015going}, as shown in the Table \ref{table:baseline}. Layers from data to inception(4e) in Table \ref{table:baseline} corresponds to the blue CNN block in Fig.\ref{fig:framewrok}, CNNg and CNNp are inception(5a) and inception(5b), respectively. The green CONV matches the subsequent 1$\times$1 convolution. The loss layers are not shown in Table \ref{table:baseline}. The Batch Normalization Layers~\cite{ioffe2015batch} are inserted before every ReLU Layer to accelerate the convergence. We employ a Convolutional Layer and a Global Average Pooling Layer (GAP) at the end of network to let our network can fit different sizes of input images. In this work, we fix input image size as 512$\times$256.

\begin{figure}
\centering 
\includegraphics[width=1\linewidth]{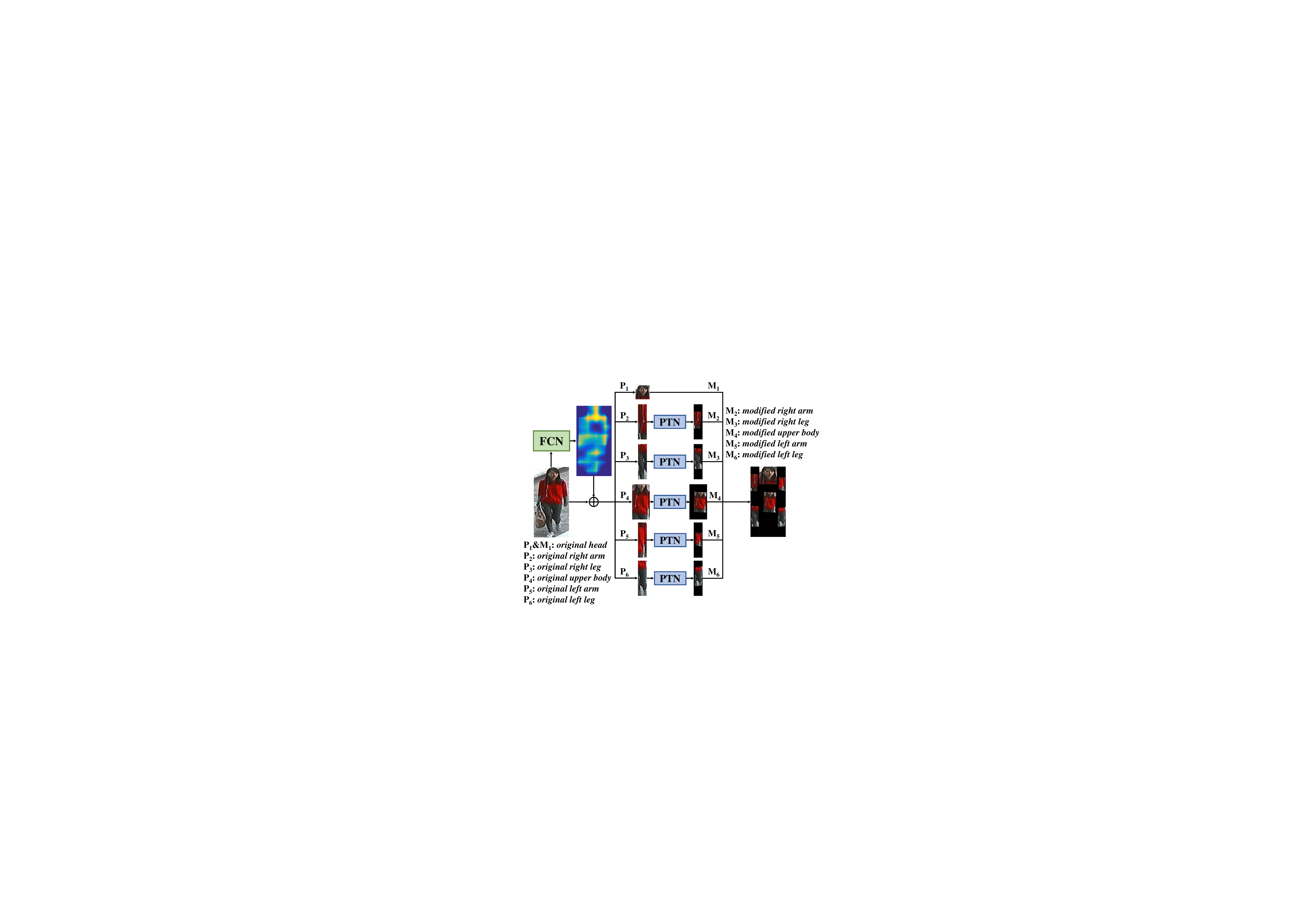}\\
\vspace{-1mm}
\caption{Illustration of Feature Embedding sub-Net (FEN). We divide the human image into 6 parts and apply an affine transformation on each part (except head part) by PTN, then we combine 6 transformed part regions together to form a new image.}
\vspace{-4mm}
\label{fig:PTN}%
\end{figure}

\subsection{Feature Embedding sub-Net }
The Feature Embedding sub-Net (FEN) is divided into four steps, including locating the joint, generating the original part images, PTN, and outputting the final modified part images.

With a given person image, FEN first locates the 14 joints of human body using human pose estimation algorithm~\cite{long2015fully}. Fig.\ref{fig:pdfes-example}(c) shows an example of the 14 joints of human body. According to number, the 14 joints are \{$head,$ $neck,$ $right shoulder,$ $right elbow,$ $right wrist,$ $left shoulder,$ $left elbow,$ $left wrist,$ $left hip,$ $left knee,$ $left ankle,$ $right hip,$ $right knee,$ $right ankle$\}. Then we propose six rectangles to cover six different parts of human body, including the head region, the upper body, two arms and two legs.

For each human joint, we calculate a response feature map $V_{i}\in\mathbb{R}^{(X,Y)}$. The horizontal and vertical dimensions of the feature maps are denoted by $X$ and $Y$, respectively. With the feature maps, the fourteen body joints $J_{i} = [X_{i},Y_{i}],(i=1,2\cdots 14)$, can be located by finding the center of mass with the feature values:
\begin{center}
\begin{equation}
J_{i} = [X_{i},Y_{i}] = [ \frac{\sum V_{i}(x_{j},y)x_{j}}{\sum V_{i}}, \frac{\sum V_{i}(x,y_{j})y_{j}}{\sum V_{i}} ],
\label{equ:centre of mass}
\end{equation}
\end{center}
where $X_{i},Y_{i}$ in Eq.\ref{equ:centre of mass} are the coordinates of joints , and $V(x,y)$ is the value of pixels in response feature maps.

Different from~\cite{zheng2017pose,zhao2017spindle} , we do not use complex pose estimation networks as the pre-trained network. Instead, we use a standard FCN~\cite{long2015fully} trained on the LSP dataset~\cite{Johnson10} and MPII human pose dataset~\cite{andriluka20142d}. In the second step, the FEN uses the 14 human joints to further locate six sub-regions (head, upper body, left arm, right arm, left leg, and right leg) as human parts. These parts are normalized through cropping, rotating, and resizing to fixed size and orientation.

As shown in Fig.\ref{fig:pdfes-example}(d), the 14 located body joints are assigned to six rectangles indicating six parts. The head part $P_{1} = [1]$, the upper body part $P_{2} = [2,3,6,9,12]$, the left arm part $P_{3} = [6,7,8]$, the right arm part $P_{4} = [3,4,5]$, the left leg part $P_{5} = [9,10,11]$, and the right leg part $P_{6} = [12,13,14]$, respectively.

For each body part set $P_{i} \in \{P_{1}, P_{2}, P_{3}, P_{4}, P_{5}, P_{6}\}$, The corresponding sub-region bounding box $H_{i}\in \{H_{1}, H_{2}, H_{3}, H_{4}, H_{5}, H_{6}\}$ can be obtained based on the location coordinates of all body joints in each part set:

\begin{equation}
H_{i}=
\left\{\begin{matrix}
[x-30,x+30, y-30, y+30 ],\ \ if \quad i=1\\
[x_{min}\!-\!10,x_{max}\!+\!10,y_{min}\!-\!10,y_{min}\!+\!10], \\ \quad\quad\quad\quad\quad\quad\quad\quad\quad\quad\quad if \quad i=2,3,4,5,6
\end{matrix}\right.
\label{eq:ourloss}
\end{equation}

\begin{figure}
\centering 
\includegraphics[width=0.75\linewidth]{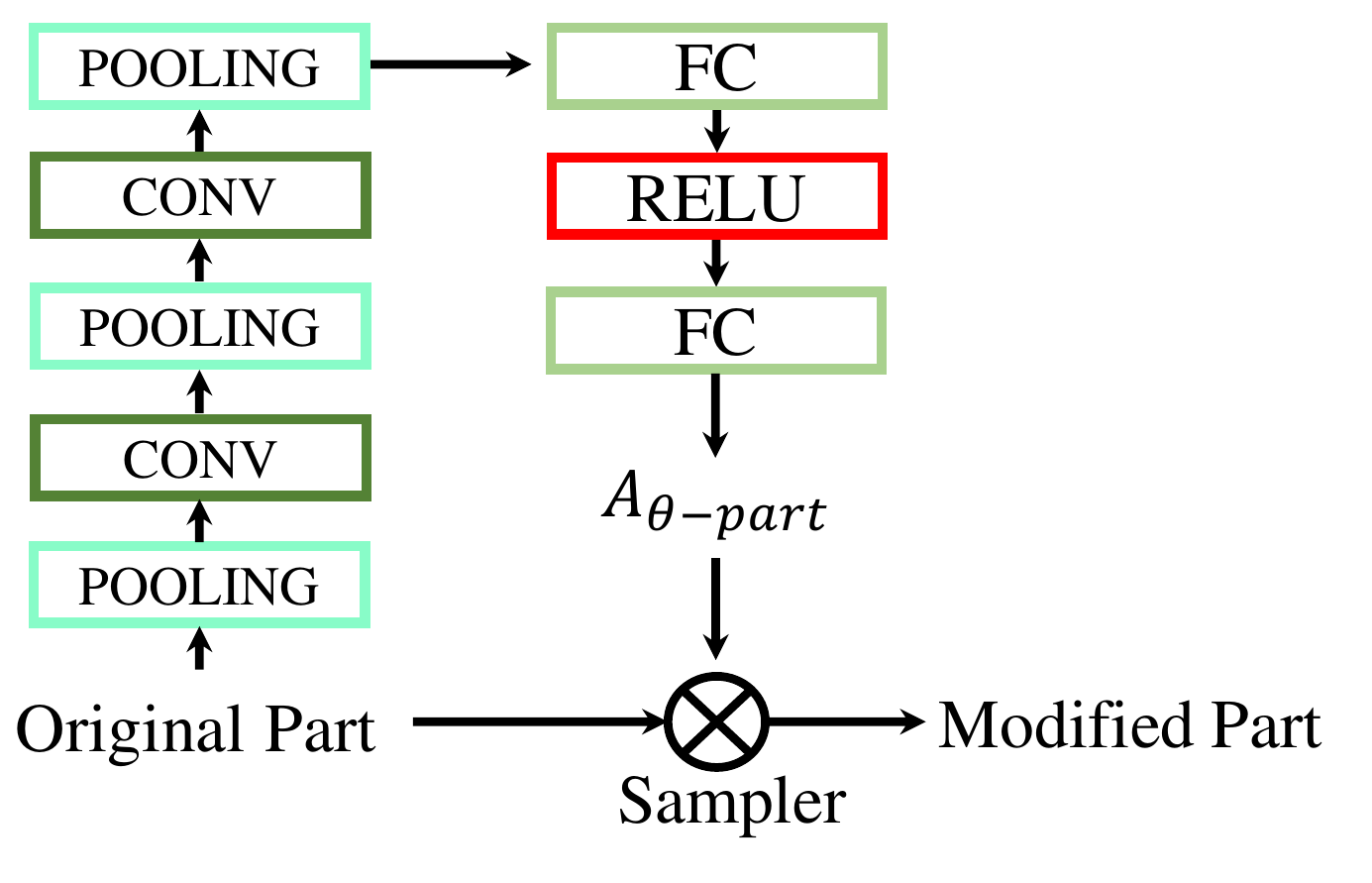}\\
\vspace{-3mm}
\caption{Detailed structure of the PTN subnet.}
\vspace{-4mm}
\label{fig:PTN1}%
\end{figure}

An example of the extracted six body sub-regions are visualized in Fig.\ref{fig:pdfes-example}(d). As shown in Fig.\ref{fig:pdfes-example}(e), these body sub-regions are normalized through cropping, rotating, and resizing to fixed sizes and orientations. All body parts are rotated to fixed vertical direction. Arms and legs are resized to 256$\times$64, upper body is resized to 256$\times$128 and head is resized to 128$\times$128. Those resized and rotated parts are combined to form the body part image. Because 6 body parts have different sizes, black area is unavoidable in body part image.

Simply resizing and rotation can not overcome the complex pose variations, especially if the pose estimations are inaccurate. We thus design a PTN modified from Spatial Transformer Networks (STN)~\cite{jaderberg2015spatial} to learn the angles required for rotating the five body parts.

STN is a spatial transformer module which can be inserted to a neural network to provide spatial transformation capabilities. It thus is potential to adjust the localizations and angles of parts. A STN is a small net which allows for end-to-end training with standard back-propagation, therefore, the introduction of STN doesn't substantially increase the complexity of training procedure. The STN consist of three components: localisation network, parameterised sampling grid, and differentiable image sampling. The localisation network takes the input feature map and outputs the parameters of the transformation. For our net, we choose affine transformation so our transformation parameter is 6-dimensional. The parameterized sampling grid computes each output pixel and the differentiable image sampling component produces the sampled output image. For more details about STN, please refer to~\cite{jaderberg2015spatial}.

As discussed above, we use a 6-dimensional parameter $A_{\theta}$ to complete affine transformation:
\begin{equation}
\begin{pmatrix}
x^{s}\\
y^{s}
\end{pmatrix}
=
A_{\theta }
\begin{pmatrix}
x^{t}\\
y^{t}\\
1
\end{pmatrix}
=
\begin{bmatrix}
 \theta _{1}&\theta _{2}&\theta _{3}\\
 \theta _{4}&\theta _{5}&\theta _{6}\\
\end{bmatrix}
\begin{pmatrix}
x^{t}\\
y^{t}\\
1
\end{pmatrix},
\label{equ:STN}%
\end{equation}
where the $\theta_{1},\theta_{2},\theta_{4},\theta_{5}$ are the scale and rotation parameters, while the $\theta_{3},\theta_{6}$ are the translation parameters. The $(x^{t},y^{t})$ in Eq.\ref{equ:STN} are the target coordinates of the output image and the $(x^{s},y^{s})$ are the source coordinates of the input image.

Usually the STN computes one affine transform for the whole image, considering a pedestrian's different parts have various orientations and sizes from each other, STN is not applicable to a part image. Inspired by STN, we design a Pose Transformer Network (PTN) which computes the affine transformation for each part in part image individually and combines 6 transformed parts together. Similar to STN, our PTN is also a small net and doesn't substantially increase the complexity of our training procedure. As a consequence, PTN has potential to perform better than STN for person images. Fig.\ref{fig:PTN} shows the detailed structure of PTN. Considering a pedestrian's head seldom has a large rotation angle, we don't insert a PTN net for the pedestrian's head part. Therefore, we totally have 5 independent PTN, namely $A_{\theta-larm}$, $A_{\theta-rarm}$, $A_{\theta-upperbody}$, $A_{\theta-lleg}$, $A_{\theta-rleg}$. Each PTN can generate a 6-dimensional transformation parameter $A_{\theta i}$ and use $A_{\theta i}$ to adjust pedestrian's part $P_{i}$, we can get modified body part $M_{i}$. By combining the five transformed parts and a head part together, we obtain the modified part image.

\begin{figure}[t]
\centering 
\includegraphics[width=0.9\linewidth]{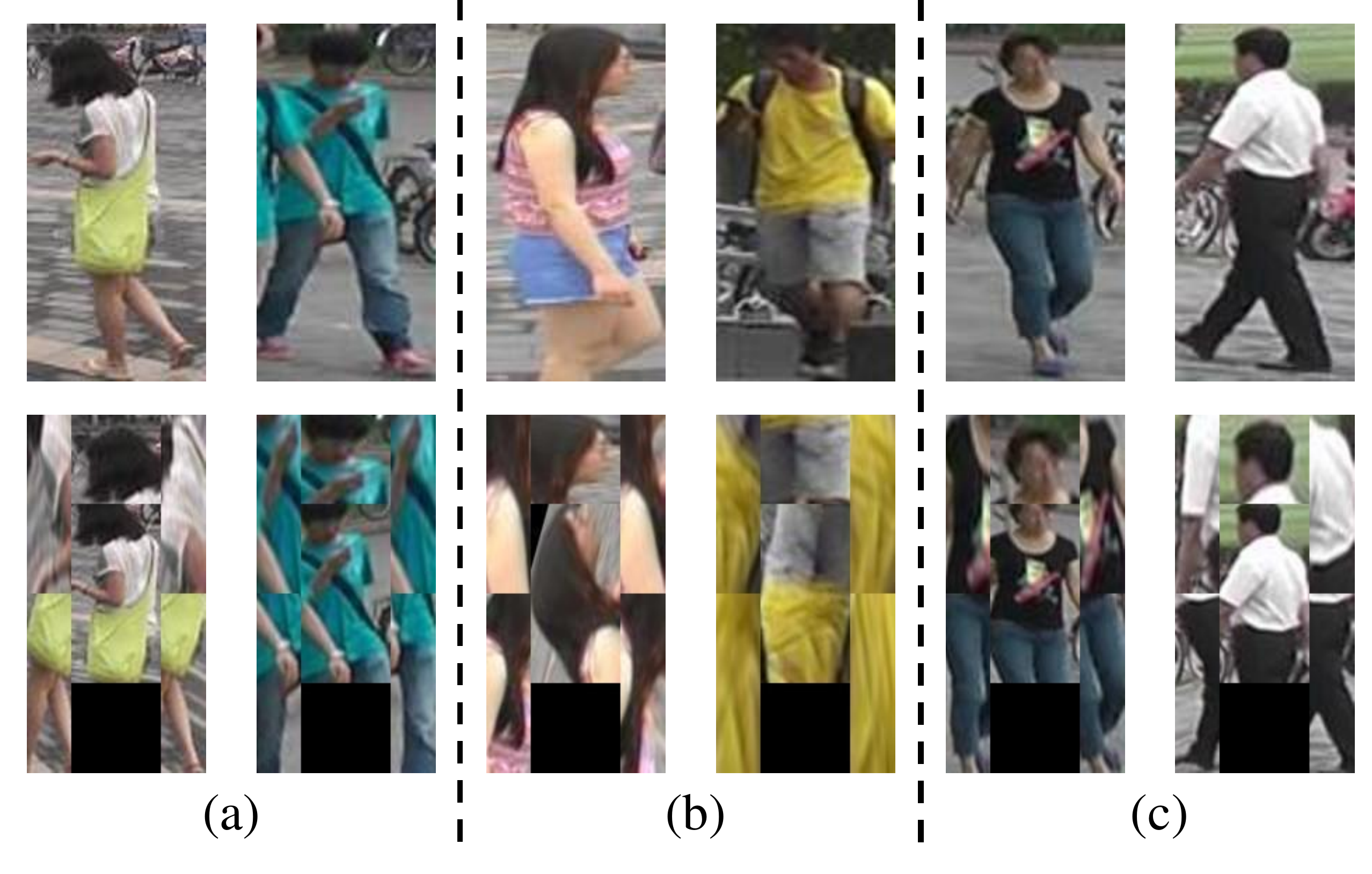}\\
\vspace{-2mm}
\caption{ Illustration of some inaccurate part detection result. (a) Arms are obscured by upper bodies. (b) Upper bodies with large variation. (c) Miss detection on arms.}
\vspace{-4mm}
\label{fig:falsepart}
\end{figure}

\subsection{Feature Weighting sub-Net}
The generated part features are combined with the global feature to generate a robust feature representation for precise person re-identification. As the poses generated by the pose detector might be affected by factors like occlusions, pose changes, etc. Then inaccurate part detection results could be obtained. Examples are shown in Fig.\ref{fig:falsepart}. Therefore, the part features could be not reliable enough. This happens frequently in real applications with unconstrained video gathering environment. Simply fusing global feature and the part feature may introduces noises. This motivates us to introduce Feature Weighting sub-Net (FWN) to seek a more optimal feature fusion. FWN is consisted with a Weight Layer and a nonlinear transformation, which decides the importance of each dimension in the part feature vector. Considering that a single linear Weight Layer might cause excessive response on some specific dimensions of the part vector, we add a nonlinear function to equalize the response of part feature vector, and the fused feature representation is
\begin{equation}
F_{fusion} = [F_{global}, tanh(F_{part}\odot W + B)],
\label{equ:aggregated feature}
\end{equation}
where the $F_{global}$ and the $F_{part}$ are the global and part feature vectors.
The $W$ and $B$ in Eq. \ref{equ:aggregated feature} are the weight and bias vectors which have the same dimensions with $F_{part}$.
The $\odot$ means the Hadamard product of two vectors, and the $[,]$ means concatenation of two vectors together. The $tanh(x) = \frac{e^{x}-e^{-x}}{e^{x}+e^{-x}}$ imposes the hyperbolic tangent nonlinearity.
$F_{fusion}$ is our final person feature generated by $F_{global}$ and $F_{part}$.

To allow back-propagation of the loss through the FWN, we give the gradient formula:
\begin{equation}
\frac{\partial f_{i}}{\partial g_{j}}=
\left\{\begin{matrix}
 1 ,\quad \text{if}\quad i=j\\
 0 ,\quad \text{if}\quad i\neq j
\end{matrix}\right.
\label{eq:diff1}
\end{equation}

\begin{equation}
\frac{\partial f_{i}}{\partial p_{k}}=
\left\{\begin{matrix}
w(1-tanh^{2}(wp_{j}+b)),~~\text{if} \quad i=k+m, \\
0 ,\quad\quad\quad\quad\quad\quad\quad\quad\quad~~~\text{if} \quad i\neq k+m.
\end{matrix}\right.
\label{eq:diff1}
\end{equation}
where $f_{i}\in F_{fusion}(i=1,2\cdots m+n)$, $g_{j}\in F_{global}(j=1,2\cdots m)$, $p_{k}\in F_{part}(k=1,2\cdots n)$,
$w_{k}\in W(k=1,2\cdots n)$, $b\in B(k=1,2\cdots n)$,
 $m$ and $n$ are the dimensions of $F_{global}$ and $F_{part}$.

\begin{figure}[t]
\centering 
\includegraphics[width=1\linewidth]{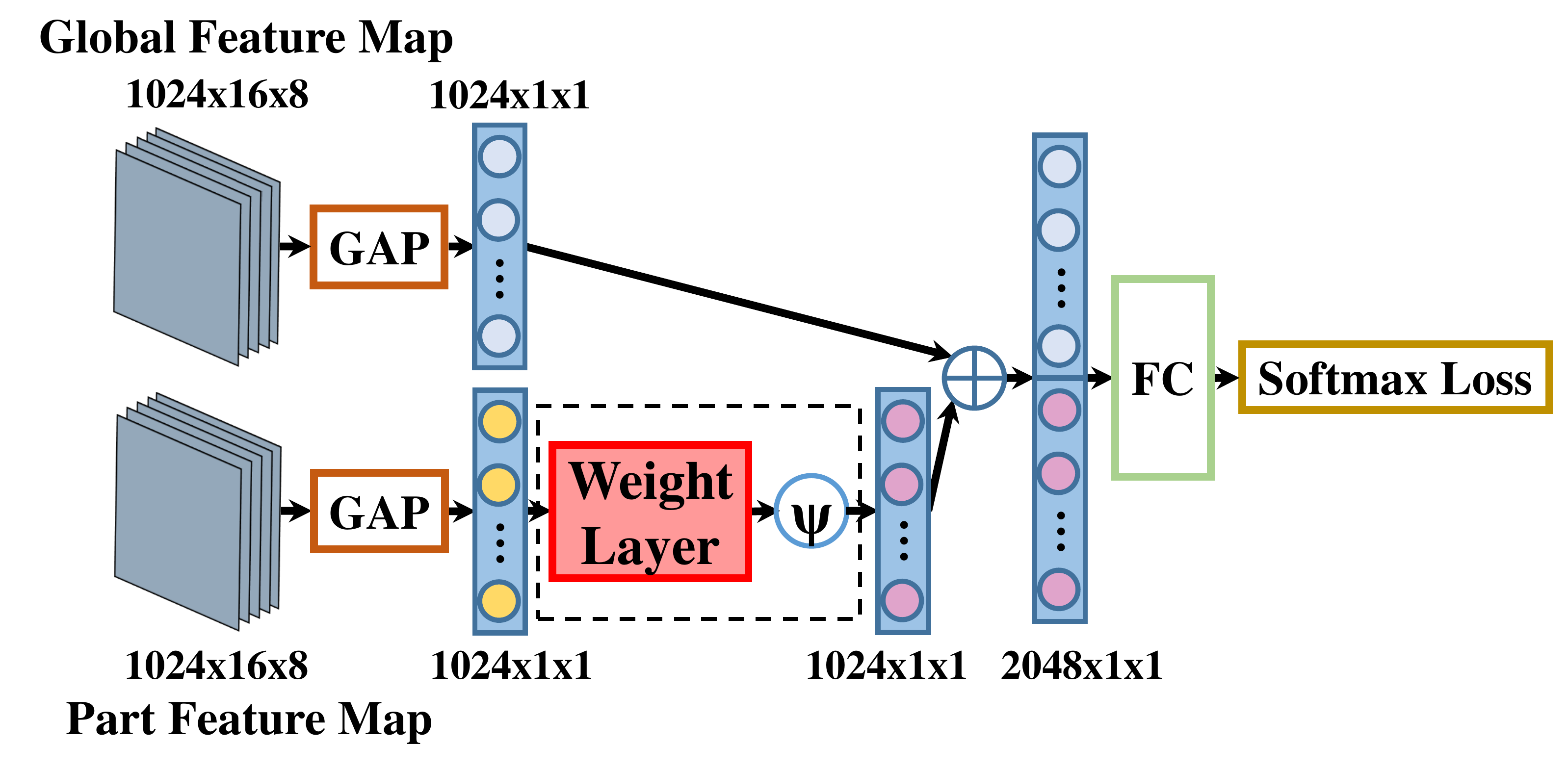}\\
\vspace{-2mm}
\caption{ Illustration of the Feature Weighting sub-Net(FWN).}
\vspace{-5mm}
\label{fig:PDFWS}%
\end{figure}

\subsection{ReID Feature Extraction}
The global feature and body-part features are learned by training the Pose-driven Deep Convolutional model. These two types of features are then fused under a unified framework for multi-class person identification. PDC extracts the global feature maps from the global body-based representation and learns a 1024-dimensional feature embedding. Similarly, a 1024-dimension feature is acquired from the modified part image after the FEN. The global body feature and the local body part features are compensated into a 2048-dimensional feature as the final representation. After being weighted by FWN, the final representation is used for Person ReID with Euclidean distance.

\section{Experiment}
\subsection{Datasets}
\label{sec:datsets}
We select three widely used person ReID datasets as our evaluation protocols, including the \emph{CUHK 03}~\cite{li2014deepreid}, \emph{Market 1501}~\cite{zheng2015scalable}, and \emph{VIPeR}~\cite{gray2007evaluating}. Note that, because the amount of images in \emph{VIPeR} is not enough for training a deep model, we combine the training sets of \emph{VIPeR}, \emph{CUHK 03} and \emph{Market 1501} together to train the model for \emph{VIPeR}.

{\emph{CUHK 03}}:
This dataset is made up of 14,096 images of 1,467 different persons taken by six campus cameras. Each person only appears in two views. This dataset provides two types of annotations, including manually labelled pedestrian bounding boxes and bounding boxes automatically detected by the Deformable-Part-Model (DPM)~\cite{felzenszwalb2008discriminatively} detector. We denote the two corresponding subsets as labeled dataset and detected dataset, respectively. The dataset also provides 20 test sets, each includes 100 identities. We select the first set and use 100 identities for testing and the rest 1,367 identities for training. We report the averaged performance after repeating the experiments for 20 times.

{\emph{Market 1501}:} This dataset is made up of 32,368 pedestrian images taken by six manually configured cameras. It has 1,501 different persons in it. On average, there are 3.6 images for each person captured from each angle. The images can be classified into two types, \ie, cropped images and images of pedestrians automatically detected by the DPM~\cite{felzenszwalb2008discriminatively}. Because \emph{Market 1501} has provided the training set and testing set, we use images in the training set for training our PDC network and follow the protocol~\cite{zheng2015scalable} to report the ReID performance.

{\emph{VIPeR}:} This dataset is made up of 632 person images captured from two views. Each pair of images depicting a person are collected by different cameras with varying viewpoints and illumination conditions. Because the amount of images in \emph{VIPeR} is not enough to train the deep model, we also perform data augmentation with similar methods in existing deep learning based person ReID works. For each training image, we generate 5 augmented images around the image center by performing random 2D transformations. Finally, we combine the augmented training images of \emph{VIPeR}, training images of \emph{CUHK 03} and \emph{Market 1501} together, as the final training set.

\begin{table}
\caption{The results on the \emph{CUHK 03}, \emph{Market 1501} and \emph{VIPeR} datasets by five variants of our approach and the complete PDC.}
\vspace{-3mm}
\label{table:components}
\footnotesize
\begin{center}
\begin{tabular}{|p{2.4cm}|p{0.8cm}<{\centering}|p{0.8cm}<{\centering}|p{0.5cm}<{\centering}|p{0.5cm}<{\centering}|p{0.6cm}<{\centering}|}
\hline
\multicolumn{1}{|c|}{\multirow{2}{*}{dataset}}&\multicolumn{2}{c|}{CUHK03}&\multicolumn{2}{c|}{\multirow{2}{*}{Market1501}}&\multirow{2}{*}{VIPeR}\\
\cline{2-3}
    &labeled& detected&\multicolumn{2}{c|}{}&\\
 \hline
method               &rank1      &rank1       &mAP         &rank1        &rank1\\
\hline
Global Only             &79.83	     &71.89       &52.84       &76.22        &37.97\\
Part Only            &53.73      &47.29       &31.74       &55.67        &22.78\\
Global+Part        &85.07      &76.33       &62.20       &81.74        &48.42\\
Global+Part+FEN  &87.15      &77.57       &62.58       &83.05        &50.32\\
Global+Part+FWN  &86.41      &77.62       &62.58       &82.69        &50.00\\
PDC            &{\bf88.70} &{\bf78.29}  &{\bf63.41}  &{\bf84.14}   &{\bf51.27}\\
\hline
\end{tabular}
\end{center}
\vspace{-6mm}
\end{table}

\subsection{Implementation Details}
\label{sec:implement}

The pedestrian representations are learned through multi-class classification CNN. We use the full body and body parts to learn the representations with Softmax Loss, respectively. We report rank1, rank5, rank10 and rank20 accuracy of cumulative match
curve (CMC) on the three datasets to evaluate the ReID performance.As for Market-1051, mean Average Precision (mAP) is also reported as an additional criterion to evaluate the performance.

Our model is trained and fine-tuned on Caffe~\cite{jia2014caffe}. Stochastic Gradient Descent (SGD) is used to optimize our model. Images for training are randomly divided into several batches, each of which includes 16 images. The initial learning rate is set as 0.01, and is gradually lowered after each $ 2\times10^{4}$ iterations. It should be noted that, the learning rate in part localization network is only 0.1\% of that in feature learning network. For each dataset, we train a model on its corresponding training set as the pretrained body-based model. For the overall network training, the network is initialized using pretrained body-based model. Then, we adopt the same training strategy as described above. We implement our approach with GTX TITAN X GPU, Intel i7 CPU, and 128GB memory.

All images are resized to $512\times256$. The mean value is subtracted from each channel (B, G, and R) for training the network. The images of each dataset are randomized in the process of training stage.

\subsection{Evaluation of Individual Components}
\label{sec:component}

We evaluate five variants of our approach to verify the validity of individual components in our PDC, \eg, components like Feature Embedding sub-Net (FEN) and Feature Weighting sub-Net (FWN). Comparisons on three datasets are summarized in Table~\ref{table:components}. In the table, ``Global Only'' means we train our deep model without using any part information. ``Global+Part'' denotes CNN trained through two streams without FEN and FWN. Based on ``Global+Part'', considering FEN is denoted as ``Global+Part+FEN''. Similarly, ``Global+Part+FWN'' means considering FWN. In addition, ``Part Only'' denotes only using part features. PDC considers all of these components.

From the experimental results, it can be observed that, fusing global features and part features achieves better performance than only using one of them. Compared with ``Global Only'', considering extra part cues, \ie, ``Global+Part'', largely improves the ReID performance and achieves the rank1 accuracy of 85.07\% and 76.33\% on \emph{CUHK 03} labeled and detected datasets, respectively. Moreover, using FEN and FWN further boosts the rank1 identification rate. This shows that training our model using PTN and Weight Layer gets more competitive performance on three datasets.

The above experiments shows that each of the components in our method is helpful for improving the performance. By considering all of these components, PDC exhibits the best performance.

\begin{table}
\caption{Comparisons on \emph{CUHK 03} detected dataset.}
\label{table:cuhk03det}
\footnotesize
\begin{center}
\begin{tabular}{|l|cccc|}
\hline
 Methods            &rank1    &rank5    &rank10   &rank20 \\
\hline
MLAPG~\cite{liao2015efficient} &51.15    &83.55    &92.05    &96.90\\
LOMO + XQDA~\cite{liao2015person} &46.25    &78.90    &88.55    &94.25\\
BoW+HS~\cite{zheng2015scalable}       &24.30    &-    &-    &-  \\
LDNS~\cite{zhang2016learning}     &54.70    &84.75    &94.80    &95.20\\
GOG~\cite{matsukawa2016hierarchical} &65.50 &88.40 &93.70 &- \\
\hline
IDLA~\cite{ahmed5improved}  &44.96    &76.01    &84.37    &93.15\\
SI+CI~\cite{wangjoint}  &52.17    &84.30    &92.30    &95.00\\
LSTM S-CNN~\cite{varior2016siamese}      &57.30     &80.10    &88.30    &-    \\
Gate S-CNN~\cite{varior2016gated} &61.80    &80.90    &88.30    &-\\
 EDM~\cite{shi2016embedding}      &52.09   &82.87   &91.78&97.17\\
\hline
PIE~\cite{zheng2017pose}    &67.10   &92.20   &96.60 &98.10\\
\hline
PDC       &{\bf78.29}   &{\bf94.83}   &{\bf97.15}   &{\bf98.43}\\
\hline
\end{tabular}
\end{center}
\vspace{-7mm}
\end{table}

\begin{table}
\caption{Comparisons on \emph{CUHK 03} labeled dataset.}
\label{table:cuhk03lab}
\footnotesize
\begin{center}
\begin{tabular}{|l|cccc|}
\hline
 Methods            &rank1    &rank5    &rank10   &rank20 \\
\hline
MLAPG~\cite{liao2015efficient} &57.96    &87.09    &94.74    &96.90\\
LOMO + XQDA~\cite{liao2015person} &52.20    &82.23    &94.14    &96.25\\
WARCA~\cite{jose2016scalable}    &78.40    &94.60    &-    &-\\
LDNS~\cite{zhang2016learning}     &62.55    &90.05    &94.80    &98.10\\
GOG~\cite{matsukawa2016hierarchical} &67.30 &91.00 &96.00 &- \\
\hline
IDLA~\cite{ahmed5improved}  &54.74   &86.50    &93.88    &98.10\\
PersonNet~\cite{wu2016personnet}  &64.80   &89.40    &94.90   &98.20\\
DGDropout~\cite{xiao2016learning}               &72.58    &91.59    &95.21    &97.72\\
EDM~\cite{shi2016embedding}                &61.32    &88.90    &96.44    &99.94\\
\hline
Spindle~\cite{zhao2017spindle}   &88.50  &97.80   &98.60 &99.20\\
\hline
PDC      &{\bf88.70}&{\bf98.61}&{\bf99.24}&{\bf99.67}\\
\hline
\end{tabular}
\end{center}
\vspace{-7mm}
\end{table}

\subsection{Comparison with Related Works}

{ \emph{CUHK 03}:} For the \emph{CUHK 03} dataset, we compare our PDC with some recent methods, including distance metric learning methods: MLAPG~\cite{liao2015efficient}, LOMO + XQDA~\cite{liao2015person}, BoW+HS~\cite{zheng2015scalable}, WARCA~\cite{jose2016scalable}, LDNS~\cite{zhang2016learning}, feature extraction method: GOG~\cite{matsukawa2016hierarchical} and deep learning based methods: IDLA~\cite{ahmed5improved}, PersonNet~\cite{wu2016personnet}, DGDropout~\cite{xiao2016learning}, SI+CI~\cite{wangjoint}, Gate S-CNN~\cite{varior2016gated}, LSTM S-CNN~\cite{varior2016siamese}, EDM~\cite{shi2016embedding},
PIE~\cite{zheng2017pose} and Spindle~\cite{zhao2017spindle}. We conduct experiments on both the detected dataset and the labeled dataset. Experimental results are presented in Table \ref{table:cuhk03det} and Table \ref{table:cuhk03lab}.

Experimental results show that our approach outperforms all distance metric learning methods by a large margin.
It can be seen that PIE~\cite{zheng2017pose}, Spindle~\cite{zhao2017spindle} and our PDC which all use the human pose cues achieve better performance than the other methods. This shows the advantages of considering extra pose cues in person ReID.
It is also clear that, our PDC achieves the rank1 accuracy of 78.29$\%$ and 88.70$\%$ on detected and labeled datasets, respectively. This leads to 11.19$\%$ and 0.20$\%$ performance gains over the reported performance of PIE~\cite{zheng2017pose} and Spindle~\cite{zhao2017spindle}, respectively.

\begin{table}
\caption{Comparison with state of the art on {Market 1501}.}
\vspace{-2mm}
\label{table:market}
\footnotesize
\begin{center}
\begin{tabular}{|l|ccccc|}
\hline
 Methods         &mAP      &rank1    &rank5    &rank10   &rank20 \\
\hline
LOMO + XQDA~\cite{liao2015person} &22.22    &43.79   &-    &- &- \\
BoW+Kissme~\cite{zheng2015scalable}       &20.76    &44.42    &63.90    &72.18    &78.95    \\
WARCA~\cite{jose2016scalable}   &-    &45.16    &68.12    &76.00    &84.00      \\
TMA~\cite{martinel2016temporal}&22.31&47.92&-&-&- \\
LDNS~\cite{zhang2016learning}       &29.87    &55.43    &-    &-    &-\\
HVIL~\cite{wang2016human}&-&78.00&-&-&-\\

\hline
PersonNet~\cite{wu2016personnet}      &26.35    &37.21   &-   &-&-\\
DGDropout~\cite{xiao2016learning}     &31.94    &59.53    &-    &-    &-\\
Gate S-CNN~\cite{varior2016gated}     &39.55    &65.88   &-   &-&-\\
LSTM S-CNN~\cite{varior2016siamese}   &35.30    &61.60    &-    &-    &-\\
\hline
PIE~\cite{zheng2017pose}  &55.95   &79.33   &90.76   &94.41   &96.65\\
Spindle~\cite{zhao2017spindle}   &-   &76.90  &91.50   &94.60   &96.70\\
\hline
PDC     &{\bf63.41}   &{\bf84.14}   &{\bf92.73}   &{\bf94.92}   &{\bf96.82}\\
\hline
\end{tabular}
\end{center}
\vspace{-6mm}
\end{table}

{\emph{Market 1501}:}
On \emph{Market 1501}, the compared works that learn distance metrics for person ReID include LOMO + XQDA~\cite{liao2015person}, BoW+Kissme~\cite{zheng2015scalable}, WARCA~\cite{jose2016scalable}, LDNS~\cite{zhang2016learning}, TMA~\cite{martinel2016temporal} and HVIL~\cite{wang2016human}. Compared works based on deep learning are
PersonNet~\cite{wu2016personnet}, Gate S-CNN~\cite{varior2016gated}, LSTM S-CNN~\cite{varior2016siamese}, PIE~\cite{zheng2017pose} and Spindle~\cite{zhao2017spindle}. DGDropout~\cite{xiao2016learning} does not report performance on Market1501. So we implemented DGDroput and show experimental results in Table~\ref{table:market}.

It is clear that our method outperforms these compared works by a large margin. Specifically, PDC achieves rank1 accuracy of 84.14\%, and mAP of 63.41\% using the single query mode. They are higher than the rank1 accuracy and mAP of PIE~\cite{zheng2017pose}, which performs best among the compared works. This is because our PDC not only learns pose invariant features with FEN but also learns better fusion strategy with FWN to emphasize the more discriminative features.

{\emph{VIPeR}:} We also evaluate our method by comparing it with several existing methods on \emph{VIPeR}. The compared methods include distance metric learning ones: MLAPG~\cite{liao2015efficient}, LOMO + XQDA~\cite{liao2015person}, BoW~\cite{zheng2015scalable}, WARCA~\cite{jose2016scalable} and LDNS~\cite{zhang2016learning}, and deep learning based ones: IDLA~\cite{ahmed5improved}, DGDropout~\cite{xiao2016learning}, SI+CI~\cite{wangjoint}, Gate S-CNN~\cite{varior2016gated}, LSTM S-CNN~\cite{varior2016siamese}, MTL-LORAE~\cite{su2017multi} and Spindle~\cite{zhao2017spindle}.

From the results shown in Table \ref{table:viper}, our PDC achieves the rank1 accuracy of 51.27\%. This outperforms most of compared methods except Spindle~\cite{zhao2017spindle} which also considers the human pose cues. We assume the reason might be because, Spindle~\cite{zhao2017spindle} involves more training sets to learn the model for \emph{VIPeR}. Therefore, the training set of Spindle~\cite{zhao2017spindle} is larger than ours, \ie, the combination of \emph{Market 1501}, \emph{CUHK03} and \emph{VIPeR}. For the other two datasets, our PDC achieves better performance than Spindle~\cite{zhao2017spindle}.

\begin{table}
\caption{Comparison with state of the art on VIPeR dataset.}
\vspace{-2mm}
\footnotesize
\begin{center}
\begin{tabular}{|l|cccc|}
\hline
 Methods            &rank1    &rank5    &rank10   &rank20 \\
\hline
MLAPG~\cite{liao2015efficient}       &40.73     &-    &82.34    &{\bf 92.37}    \\
LOMO + XQDA~\cite{liao2015person}            &40.00   &67.40    &80.51    &91.08\\
BoW~\cite{zheng2015scalable}       &21.74    &-    &-    &-  \\
WARCA~\cite{jose2016scalable}        &40.22    &68.16    &80.70    &91.14\\
LDNS~\cite{zhang2016learning}    &42.28 &71.46   &82.94&92.06\\
\hline
IDLA~\cite{ahmed5improved}  &34.81    &76.12    &-    &-\\
DGDropout~\cite{xiao2016learning}   & 38.6 & -& - & -\\
SI+CI~\cite{wangjoint}  &35.80   &67.40    &83.50    &-\\
LSTM S-CNN~\cite{varior2016siamese}  &42.40 &68.70 &79.40&-\\
Gate S-CNN~\cite{varior2016gated} &37.80    &66.90    &77.40    &-\\
MTL-LORAE~\cite{su2017multi}&42.30&72.20&81.60&89.60\\
\hline
Spindle~\cite{zhao2017spindle}   &{\bf 53.80}  &{\bf 74.10}   &83.20 &92.10\\
\hline
PDC     &51.27&74.05&{\bf 84.18}&91.46\\
\hline
\end{tabular}
\end{center}
\vspace{-5mm}
\label{table:viper}
\end{table}

\subsection{Evaluation of Feature Weighting sub-Net}
\label{sec:weight}
To test the effectiveness of Feature Weighting sub-Net (FWN), we verify the performance of five variants of FWN, which are denoted as $W_{k}$, $k$ = \{0,1,2,3,4\}, where $k$ is the number of Weight Layers in FWN with nonlinear transformation. For example, $W_{2}$ means we cascade two Weight Layers with nonlinear transformation, $W_{0}$ means we only have one Weight Layer without nonlinear transformation.

\begin{table}
\caption{Performance of five variants of FWN on \emph{CUHK 03}, \emph{Market 1501} and \emph{VIPeR}, respectively.}
\vspace{-2mm}
\footnotesize
\begin{center}
\begin{tabular}{|p{1.5cm}<{\centering}|p{0.8cm}<{\centering}|p{0.8cm}<{\centering}|p{0.8cm}<{\centering}|p{0.8cm}<{\centering}|p{0.8cm}<{\centering}|}
\hline
\multirow{2}{*}{dataset}& \multicolumn{2}{c|}{CUHK03}&\multicolumn{2}{c|}{\multirow{2}{*}{Market1501}}&\multirow{2}{*}{VIPeR}\\
\cline{2-3}
           &labeled      &detected    &\multicolumn{2}{c|}{}&\\
 \hline
type &rank1        &rank1       &mAP         &rank1        &rank1\\
\hline
$W_{0}$    &88.18	     &77.58       &62.58       &83.05        &42.09\\
$W_{1}$    &{\bf 88.70}  &{\bf78.29}  &{\bf63.41}  &{\bf84.14}   &{\bf43.04}\\
$W_{2}$    &88.14        &77.48       &62.20       &82.72        &41.77\\
$W_{3}$    &87.97        &77.29       &61.99       &82.48        &41.77\\
$W_{4}$    &87.69        &77.17       &61.67       &82.42        &41.14\\
\hline
\end{tabular}
\end{center}
\vspace{-6mm}
\label{table:weight}
\end{table}

\begin{figure}[t]
\centering 
\includegraphics[width=0.9\linewidth]{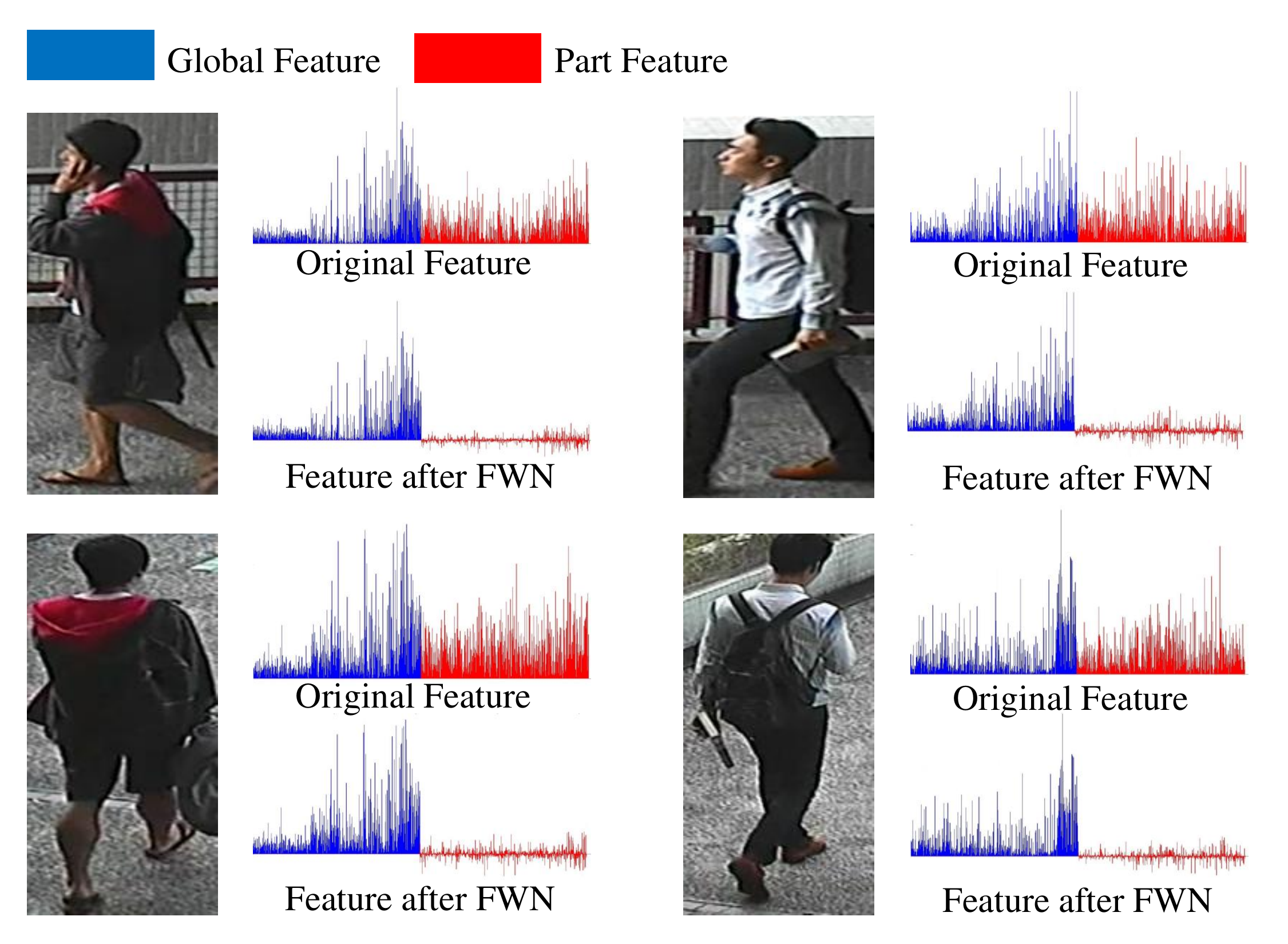}\\
\caption{Examples of fused features before and after Feature Weighting sub-Net (FWN). The two images on the left side contains the same person. The other two images contains another person. FWN effectively keeps the discriminative feature and suppresses the noisy feature.}
\label{fig:weight}
\vspace{-5mm}
\end{figure}

The experimental results are shown in Table \ref{table:weight}. As we can see that one Weight Layer with nonlinear transformation gets the best performance on the three datasets. The ReID performance starts to drop as we increase of the number of Weight Layers, despite more computations are being brought in. It also can be observed that, using one layer with nonlinear transformation gets better performance than one layer without nonlinear transformation, \ie, $W_{0}$. This means adding one nonlinear transformation after a Weight Layer learns more reliable weights for feature fusion and matching. Based on the above observations, we adopt $W_{1}$ as our final model in this paper. Examples of features before and after FWN are shown Fig.~\ref{fig:weight}.

\section{Conclusions}
\label{sec_conclusion}
This paper presents a pose-driven deep convolutional model for the person ReID. The proposed deep architecture explicitly leverages the human part cues to learn effective feature representations and adaptive similarity measurements. For the feature representations, both global human body and local body parts are transformed to a normalized and homologous state for better feature embedding. For similarity measurements, weights of feature representations from human body and different body parts are learned to adaptively chase a more discriminative feature fusion. Experimental results on three benchmark datasets demonstrate the superiority of the proposed model over current state-of-the-art methods.

~\\
\begin{small}
\textbf{Acknowledgments}
This work is partly supported by National Science Foundation of China under Grant No. 61572050, 91538111, 61620106009, 61429201, 61672519, and the National 1000 Youth Talents Plan. Dr. Qi Tian is supported by ARO grant W911NF-15-1-0290 and Faculty Research Gift Awards by NEC Laboratories of America and Blippar.
\end{small}

{\small
\bibliographystyle{ieee}
\bibliography{egbib}
}

\end{document}